# MedGemma vs GPT-4: Open-Source and Proprietary Zero-shot Medical Disease Classification from Images

Md. Sazzadul Islam Prottasha[1] *, Nabil Walid Rafi[2]
[1, 2] Department of Information and Communication Technology,
Bangladesh University of Professionals, Dhaka, Bangladesh
[1] prottasha@bup.edu.bd, [2] nabilwalidrafi@gmail.com
*Corresponding Author

***Abstract*— Multimodal Large Language Models (LLMs) introduce an emerging paradigm for medical imaging by interpreting scans through the lens of extensive clinical knowledge, offering a transformative approach to disease classification. This study presents a critical comparison between two fundamentally different AI architectures: the specialized open-source agent MedGemma and the proprietary large multimodal model GPT-4 for diagnosing six different diseases. The MedGemma-4b-it model, fine-tuned using Low-Rank Adaptation (LoRA), demonstrated superior diagnostic capability by achieving a mean test accuracy of 80.37% compared to 69.56% for the untuned GPT-4. Furthermore, MedGemma exhibited notably higher sensitivity in high-stakes clinical tasks, such as cancer and pneumonia detection. Quantitative analysis via confusion matrices and classification reports provides comprehensive insights into model performance across all categories. These results emphasize that domain-specific fine-tuning is essential for minimizing hallucinations in clinical implementation, positioning MedGemma as a sophisticated tool for complex, evidence-based medical reasoning.**

**Keywords— MedGemma; GPT-4; disease classification; medical diagnostics; artificial intelligence**

## I. INTRODUCTION

Diabetes affects approximately 589 million adults (20-79 years) worldwide [1]. Key examples underscore this severity: Breast cancer sees 2.3 million new cases annually [2]. Pneumonia accounts for 14% of all under 5 deaths [3]. Alzheimer's or other dementia diseases impact over 55 million globally [4]. Cardiovascular diseases caused 19.8 million deaths in 2022 [5], and chronic kidney disease affects over 674 million people globally [6]. These numbers illustrate the severity of chronic diseases and the increasing need to handle them. While traditional deep learning models demonstrated excellent disease classification accuracy, they often function as black boxes that offer high accuracy but limited transparency. These models frequently fail to provide the reasoning behind a diagnosis, making it difficult for clinicians to trust or verify their outputs without manual review. AI-powered disease classification uses convolutional neural networks (CNNs) and other deep learning techniques to analyze images for diagnostic purposes. The earlier works often neglected textual clinical data and multimodal integration, which limits their applicability to diverse datasets [7]. In contrast, Modern Large Language Models (LLMs) and specialized multimodal frameworks bridge this gap by seamlessly integrating textual clinical records with medical imaging, providing a comprehensive diagnostic context. LLMs' potential has been studied in medical question answering and diagnostic performance. The LLMs have shown insufficiency in handling complex medical terminology when deployed in the real world [8]-[10]. General-purpose LLMs like GPT-4 struggle with personalized medicine and genomic data integration [11]. Moreover, general LLMs like GPT-4 pose challenges in transparency and misinformation risks when integrated into clinical workflows [12]. Since the General-purpose LLMs are not specifically trained on medical data, gaps are also addressed in conversational diagnostics, medical care, and standardized reporting [13]. This lack of specialized training makes it difficult for general-purpose models to provide the clear, evidence-based reasoning that healthcare professionals require.

This study overcomes these limitations by evaluating MedGemma, a medically specialized multimodal model, against widely used general purpose LLM GPT-4 across datasets of six chronic diseases. These data are on skin cancer, Alzheimer's disease, breast cancer, cardiovascular disease, pneumonia, and chronic kidney disease. A comprehensive evaluation is performed and using the features of a wide variety of datasets, this study explores the potential of medically specialized model MedGemma in clinical applications.

## II. LITERATURE REVIEW

General-purpose LLMs like GPT-4 have revolutionized natural language processing tasks. They have achieved state-of-the-art performance in daily tasks such as text generation, sentiment analysis, and question answering [14], [15]. The use of GPT models for diagnosis, classification and reasoning in healthcare has also been examined [16]. GPT-4 can perform remarkably well on medical licensing exams and suggest differential diagnoses based on textual symptoms. However, they are medical generalists and can suffer from hallucinations or a lack of grounding in specific clinical guidelines [17]. GPT models are not well-suited for clinical text classification. They have a lack of interpretation of medical abbreviations [18]. LLMs like PaLM can encode clinical knowledge but their general-purpose nature limits precision in clinical diagnostic tasks. Flan-PaLM reached 67.6% on MedQA [19], and Med-PaLM 2 later reached 86.5% through medical-domain fine-tuning and ensemble refinement [9]. A systematic review and meta-analysis reported an overall diagnostic accuracy of 52.1% for generative AI models, with no significant difference from non-expert physicians but significantly lower accuracy than expert physicians [10]. It restates the limitations of general-purpose models.

Medically specialized models, such as BioBERT, ClinicalBERT, and MedGemma, have been trained to handle these limitations. They are trained on domain data such as PubMed abstracts, clinical notes, and medical images [20]-[22]. The BioBERT model is pretrained on PubMed and PMC. It excels in tasks like named entity recognition and relation extraction. ClinicalBERT is further pretrained on MIMIC-III clinical notes and improves clinical named entity recognition and natural language inference on MedNLI [21]. MedGemma is a family of open multimodal models built on Gemma 3 and trained on medical text and images, including chest X-rays, histopathology, dermatology and fundus images [22]. This study [24] investigates the integration of Explainable Artificial Intelligence (XAI) as a vital mechanism for bridging the gap between high-performance diagnostic algorithms and clinical trust. While deep learning models demonstrate superior diagnostic potential, their black-box architecture presents substantial challenges to regulatory compliance, ethical accountability, and physician engagement. A key finding is that most current research fails to involve medical experts in the design process. To fix this, the study suggests creating interactive visual dashboards that allow doctors to check and confirm AI suggestions [24]. Such a paradigm shift would transition AI systems from autonomous diagnostic tools to reliable expert companions, establishing a framework where transparency and predictive accuracy are balanced to ensure safer, evidence-based medical decision-making.

Table I shows an overview of various studies on AI and LLMs in clinical disease diagnosis application.

TABLE I. STUDY OF AI AND LLMS IN CLINICAL APPLICATIONS

| Study | Models Compared | Datasets | Methodology | Key Findings |
|---|---|---|---|---|
| [7] | CNN, RNN hybrids | ECG data | Comprehensive review | 99.93% accuracy in arrhythmia detection by Deep Learning. |
| [8] | Flan-PaLM, Med-PaLM | MultiMedQA | Few-shot, chain-of-thought and instruction prompt tuning | Flan-PaLM reaches 67.6% on MedQA; clinician review still finds gaps. |
| [9] | Med-PaLM 2, Flan-PaLM | MedQA, MedMCQA, PubMedQA | Medical fine-tuning, ensemble refinement | Med-PaLM 2 achieves 86.5% on MedQA, surpassing prior models. |
| [12] | LLMs, Physicians | MIMIC-IV (2,400 patients, 4 abdominal pathologies) | Curated dataset, clinical simulation | LLMs fail to follow guidelines and underperform compared to physicians. |
| [13] | AMIE, Physicians | Text-based consultations | Self-play environment, human evaluation | AMIE outperforms physicians on 30/32 axes. |
| [14] | GPT-3 | Diverse internet texts | General-purpose pretraining, few-shot in-context learning | Strong few-shot results on general NLP benchmarks; no medical evaluation reported. |
| [20] | BioBERT, BERT | PubMed, PMC | Pretraining on biomedical texts, fine-tuning for NER, RE | BioBERT outperforms BERT by 0.62% F1 (NER), 2.80% F1 (RE) and 12.24% MRR (QA). |
| [21] | ClinicalBERT, BERT | MIMIC-III | Further pretraining on clinical notes; evaluated on NER and MedNLI | Clinical BERT variants reach up to 82.7% accuracy on MedNLI. |
| [22] | MedGemma 4B, MedGemma 27B | Medical images and medical text | Medical adaptation of Gemma 3, instruction tuning | MedQA accuracy of 64.4% (4B) and 87.7% (27B text). |
| [24] | XAI models, DL models | Medical imaging datasets | Model-agnostic XAI for medical decision support | XAI enhances trust but lacks physician involvement in validation. |

Despite recent advancements in medical AI, there remains a critical lack of direct comparisons between MedGemma and general-purpose models like GPT-4. While general LLMs like GPT-4 are versatile, existing literature reveals a significant gap in their reasoning abilities regarding medical explainability in rare pathologies. In contrast, MedGemma built on the Gemma 3 architecture is designed to perform complex, multi-step clinical workflows rather than just question answering [22]. Unlike general-purpose models, MedGemma is trained on de-identified medical data, including chest X-rays (MIMIC-CXR), CT and MRI slices, histopathology, dermatology and ophthalmology images, and medical question-answer text [22]. Since general-purpose LLMs lack specialized training, they struggle to provide the clear, evidence-based reasoning essential for professional medical applications. While GPT-4 excels at various linguistic tasks, it lacks the deep clinical grounding found in MedGemma. Therefore, in this study a rigorous comparison has been performed to determine whether MedGemma's medical tuning leads to fewer hallucinations in critical diagnostic scenarios.

## III. Methodology

The methodology comprises the evaluation of the medical model MedGemma-4b-it and GPT-4o across a wide variety of six chronic disease datasets. The images of the dataset have been preprocessed and fed into the models for a comprehensive evaluation.

### A. Dataset Collection

To ensure a rigorous and comprehensive evaluation, six distinct disease datasets were utilized to create a broad benchmark spanning multiple clinical modalities. Incorporating such diverse range of medical data allowed the study to move beyond narrow, task-specific testing toward a fair and generalized assessment of model performance. The HAM10000 dataset [28] has 10015 images of skin cancer. The seven lesion classes are Actinic Keratoses and Intraepithelial Carcinoma (akiec), Basal Cell Carcinoma (bcc), Benign Keratosis Lesions (bkl), Dermatofibroma (df), Melanoma (mel), Melanocytic Nevi (nv), and Vascular Lesions (vasc). For classifying Alzheimer's Disease, the OASIS MRI dataset includes 86,437 brain MRI images from 461 patients, converted to Nifti (.nii) and JPEG formats. Images are sliced along the z-axis and classified into four categories based on Clinical Dementia Rating (CDR): non-demented, very mild demented, mild demented, and moderate demented [29]. The Curated Breast Imaging Subset of DDSM (CBIS-DDSM) contains 10,239 mammography images from 1,566 participants with 3,568 annotated lesions [30]. This study used the JPEG release of CBIS-DDSM on Kaggle [38], whose case-description files list the same 3,568 lesions (Table II). Each lesion carries a verified pathology label: malignant (1,457), benign (1,429), or benign without callback (BWC, 682), the last being benign findings that did not need a recall for further work-up. CBIS-DDSM contains no normal mammograms, so the three classes used here are malignant, benign, and BWC. In the classification of cardiovascular diseases, the ECG images dataset contains 929 12-lead ECG images across four categories: Myocardial Infarction (MI) (240), Abnormal Heartbeat (233), History of Myocardial Infarction (172), and Normal (284) [31]. Each ECG was split into its 12 single-lead images, giving 11,148 images. For pneumonia prediction, 5,856 pediatric chest X-ray images were used. They come from a single centre, Guangzhou Women and Children's Medical Center, and cover patients aged one to five years [32]. The CT Kidney Dataset includes 12,446 images in lossless JPEG format, categorized as Normal (5,077), Cyst (3,709), Tumor (2,283), and Stone (1,377), sourced from hospitals in Dhaka, Bangladesh [33]. After collecting the datasets, each dataset was verified for completeness, with metadata cross-checked against original sources to ensure data integrity.

Table II shows an overview of the datasets along with samples for each class.

TABLE II. Dataset Descriptions

| Disease | Image Quantity | Patients/ Samples | Classes |
|---|---|---|---|
| Skin Cancer [28] | 10,015 | - | AKIEC (327), BCC (514), BKL (1099), DF (115), NV (6705), MEL (1113), VASC (142) |
| Alzheimer's Disease [29] | 86,437 | 461 | Non-Demented (67222), Very Mild Dementia (13725), Mild Dementia (5002), Moderate Dementia (488) |
| Breast Cancer [30] | 3,568 | 1566 | BWC (682), Benign (1429), Malignant (1457) |
| Cardiovascular Disease [31] | 11,148 | 929 ECGs | Myocardial Infarction (MI) (2880), Historic MI (2064), Abnormal Heartbeat (2796), Normal (3408) |
| Pneumonia [32] | 5,856 | - | Normal (1583), Pneumonia (4273) |
| Chronic Kidney Disease [33] | 12,446 | - | Normal (5077), Cyst (3709), Tumor (2283), Stone (1377) |

### B. Data Preprocessing

All images were resized to 896×896 pixels, the input resolution of the MedGemma vision encoder, and processed with the same pipeline. Pixel intensity values were min-max scaled to the range [0, 1]. Data augmentation with random ±10° rotations, horizontal flips, and ±20% intensity changes was applied to the training images only; validation and test images were not augmented. MedGemma-4b-it does not need a separate feature extractor: its built-in SigLIP vision encoder, a 400M-parameter Vision Transformer [25], [26], encodes each 896×896 image into 256 image tokens that are passed to the Gemma 3 language model [22]. The images were therefore supplied to MedGemma through its Hugging Face processor together with the text prompt, as shown in Fig. 1. GPT-4o received the same preprocessed test images.

### C. Model Specification and Architecture

MedGemma is a collection of open models from Google Research and Google DeepMind, built on Gemma 3 and adapted for medical text and image comprehension [22]. MedGemma-4b-it, the instruction-tuned 4-billion-parameter multimodal variant used here, pairs the Gemma 3 4B decoder (34 layers, hidden size 2,560) with a 400M-parameter SigLIP image encoder that was further trained on de-identified medical images. Its training data cover chest X-rays (e.g., MIMIC-CXR), CT and MRI slices, histopathology patches, dermatology images, and fundus photographs, together with medical text; genomic data were not included [22]. In this study, MedGemma produces its prediction as text: the fine-tuned model answers a multiple-choice prompt with the class name, which is then parsed into a class label. Multimodal models of this type are increasingly evaluated for clinical decision support [27].

GPT-4 is a large multimodal Transformer model that accepts image and text inputs and produces text outputs [23]. It succeeds GPT-3, which has 175 billion parameters, 96 layers, 96 attention heads, and a hidden size of 12,288 [14]. This study used GPT-4o, the natively multimodal version of GPT-4 released by OpenAI in May 2024 [36]; throughout this paper, GPT-4 refers to GPT-4o. OpenAI has not disclosed the architecture, parameter count, training data, or image encoder design of GPT-4 or GPT-4o [23], [36].

In this research, MedGemma-4b-it model has been used and fine-tuned on the training dataset. GPT-4o was evaluated zero-shot and was not fine-tuned. Its prompt was "Classify the images of {Disease_Name} having {Class_names}" (Fig. 1), where {Class_names} lists the class names of that dataset, the same class list given to MedGemma.

### D. MedGemma Model Training

The MedGemma-4b-it model has been trained on an NVIDIA A100 GPU cluster (8 GPUs, 40GB). Fine-tuning followed the official MedGemma Hugging Face notebook [37]. Each training example was formatted as a two-turn conversation: the user turn holds the image and a multiple-choice prompt listing the class names of that dataset, and the assistant turn holds the correct class name. The model was trained with the standard next-token cross-entropy loss of supervised fine-tuning, with image and padding tokens masked from the loss. To fit GPU memory, the base model was loaded with 4-bit NF4 quantization and double quantization, with bfloat16 compute (QLoRA). Low Rank Adaptation (LoRA) was utilized to train the model at tractable parameters. For GPT-3 175B, LoRA reduced the number of trainable parameters by 10,000 times and GPU memory by 3 times [34]. Table III shows the hyperparameters used to train MedGemma-4b-it model using LoRA.

TABLE III. MEDGEMMA TRAINING PARAMETERS

| Parameter | Value |
|---|---|
| Optimizer | AdamW (fused) |
| Quantization | 4-bit NF4 (QLoRA) |
| Learning Rate | $2\times10^{-5}$, linear, 3% warm-up |
| Batch Size | 32 |
| Epochs | 15 |
| LoRA rank r / alpha | 16 / 16 |
| LoRA dropout | 0.05 |
| Gradient Clipping Norm | 0.3 |

LoRA adapters (r = 16, α = 16, dropout 0.05) were attached to all linear layers of the model, and the language-model head and token embeddings were also trained, as in the notebook [37]. Training used the TRL SFTTrainer with the fused AdamW optimizer [35], bfloat16 precision, gradient checkpointing, a linear learning-rate schedule with 3% warm-up, and gradient clipping at a norm of 0.3. A grid search over learning rates of $1\times10^{-5}$, $2\times10^{-5}$ and $5\times10^{-5}$ and batch sizes of 16, 32 and 64 selected a learning rate of $2\times10^{-5}$ and a batch size of 32 by validation performance. The model was trained for 15 epochs, and TensorBoard was used to track the loss and accuracy curves. At test time, the fine-tuned model generated its answer with greedy decoding (at most 40 new tokens), and the generated text was matched to a class name; a response that matched no class was counted as an error. Fig. 1 illustrates the workflow of this research.

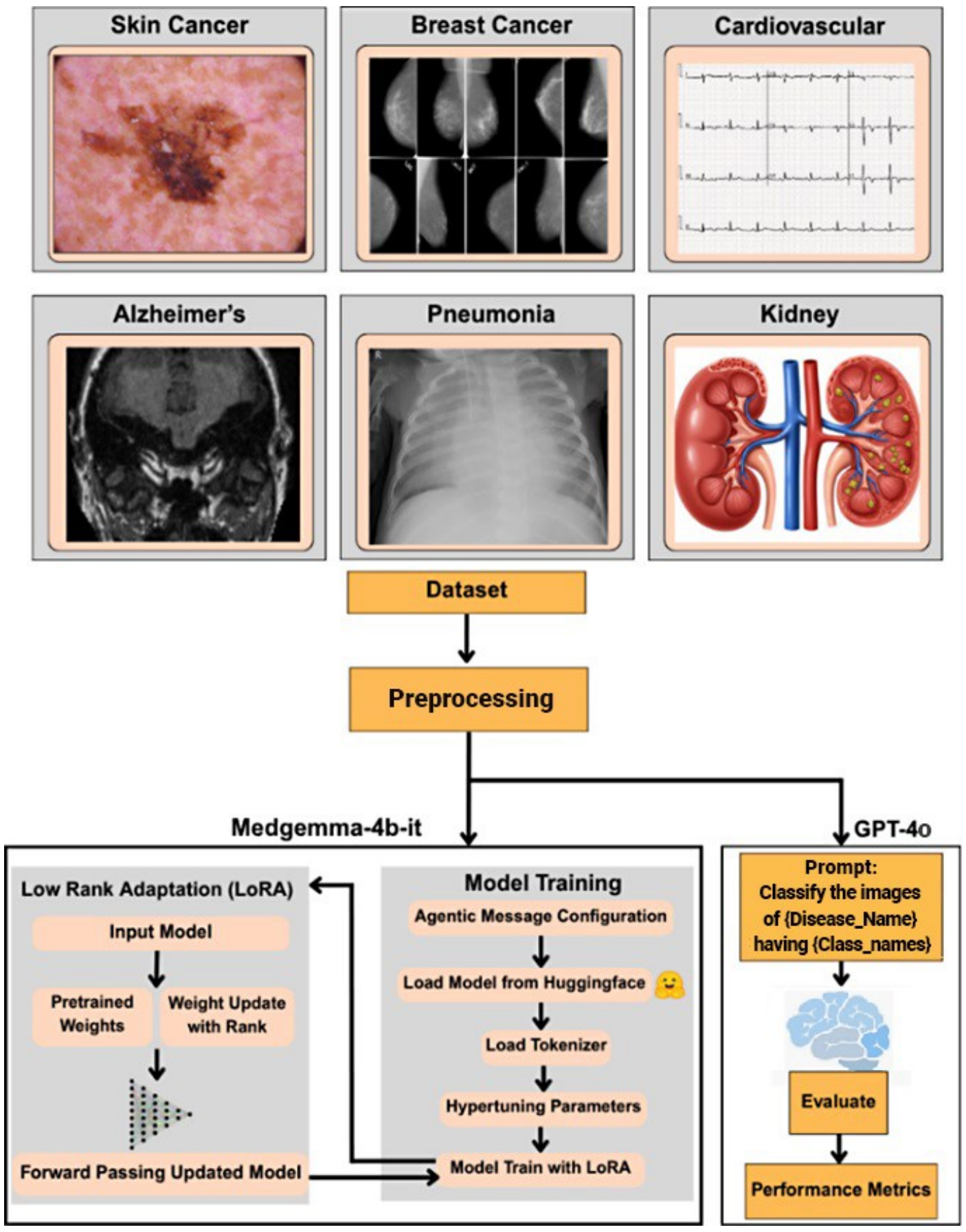

Fig. 1. Flow chart of disease classification

## IV. RESULT

MedGemma and GPT-4o's performance was assessed on 6 different disease datasets: skin cancer, Alzheimer's disease, breast cancer, cardiovascular, pneumonia, and chronic kidney disease. The skin cancer, Alzheimer's disease, breast cancer, pneumonia and chronic kidney disease datasets were split at the patient level, so all images of one patient fall in a single split. The cardiovascular dataset was split at the single-lead image level: no image appears in more than one split, but leads cut from the same 12-lead ECG can fall in different splits. Because only MedGemma was trained on these images, this can raise its cardiovascular test accuracy relative to GPT-4o, and the cardiovascular gap in Table IV should be read with this in mind. For each dataset, 70% of the images were used to fine-tune MedGemma-4b-it with LoRA. A held-out test set of a different size was then drawn for each dataset to compare MedGemma with GPT-4o: 1,513 images for skin cancer (15.1% of the dataset), 2,250 for Alzheimer's disease (2.6%), 704 for breast cancer (19.7%), 1,113 for cardiovascular disease (10.0%), 585 for pneumonia (10.0%), and 1,050 for chronic kidney disease (8.4%). The remaining images formed the validation set, i.e., 14.9%, 27.4%, 10.3%, 20.0%, 20.0%, and 21.6% of the respective datasets. For Alzheimer's disease, the Moderate Dementia class has only 488 original images, so augmentation was used to expand this class in the training set only; the 300 Moderate Dementia test images are original, unaugmented scans. Fig. 2 shows the training accuracy of the MedGemma-4b-it model for each disease.

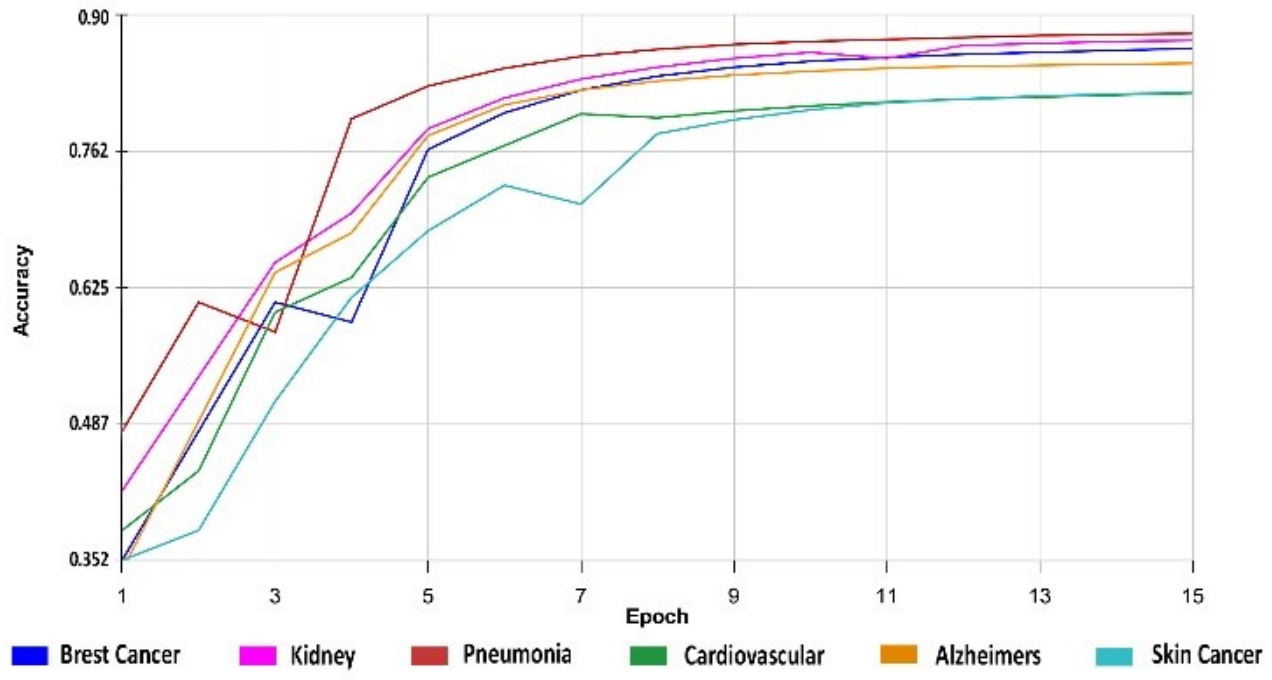

Fig. 2. Training accuracy of MedGemma-4b-it model.

After the initial training on MedGemma-4b-it model, both MedGemma and GPT-4o were benchmarked using an identical, unseen test set for each disease class. While the MedGemma model was specifically trained before testing the images, GPT-4o's performance was solely assessed with the zero-shot prompt shown in Fig. 1. Table IV demonstrates performance metrics across various datasets.

TABLE IV. PERFORMANCE COMPARISON

| Dataset | Model | Accuracy (%) | | |
|---|---|---|---|---|
| | | Train | Validation | Test |
| Skin Cancer | MedGemma | 82.2 | 80.4 | 79.05 |
| | GPT-4o | - | - | 69.40 |
| Alzheimer's Disease | MedGemma | 85.1 | 82.3 | 80.44 |
| | GPT-4o | - | - | 71.16 |
| Breast Cancer | MedGemma | 86.6 | 82.1 | 81.11 |
| | GPT-4o | - | - | 70.45 |
| Cardio vascular Disease | MedGemma | 82.1 | 80.7 | 79.34 |
| | GPT-4o | - | - | 67.65 |
| Pneumonia | MedGemma | 88.1 | 84.5 | 81.71 |
| | GPT-4o | - | - | 69.91 |
| Chronic Kidney Disease | MedGemma | 87.4 | 83.1 | 80.57 |
| | GPT-4o | - | - | 68.76 |
| **Average** | **MedGemma** | **85.25** | **82.18** | **80.37** |
| | **GPT-4o** | **-** | **-** | **69.56** |

The result reported in Table IV shows that among all the datasets, MedGemma-4b-it achieved a mean accuracy of 80.37%, outperforming GPT-4o's mean accuracy of 69.56%. The analysis of the provided dataset indicates that the MedGemma model consistently demonstrates superior diagnostic performance compared to GPT-4o across a diverse set of six medical classification tasks. The 95% Wilson confidence intervals of the two models' test accuracies do not overlap on any dataset; the closest pair is breast cancer, at 78.1-83.8% for MedGemma and 67.0-73.7% for GPT-4o.

This performance gap is uniformly observed in every category with MedGemma's highest test accuracy recorded in Pneumonia detection at 81.71% and its lowest test accuracy in Skin Cancer at 79.05%. Conversely, GPT-4o's performance ranges from a maximum of 71.16% for Alzheimer's Disease to a minimum of 67.65% for cardiovascular disease.

In skin cancer classification, both MedGemma and GPT-4o struggled to correctly classify the diseases as illustrated in the confusion matrix in Fig. 3. In Actinic Keratoses (AKIEC) class, MedGemma recorded 28 correct predictions out of 43 cases, slightly outperforming GPT-4o's 23. Both models had their lowest precision on the AKIEC and melanoma (MEL) classes; MedGemma, for example, labelled 108 NV lesions as MEL. For three NV images, GPT-4o returned no valid class label; these are counted as errors, so the NV row of Fig. 3(b) sums to 907 rather than 910.

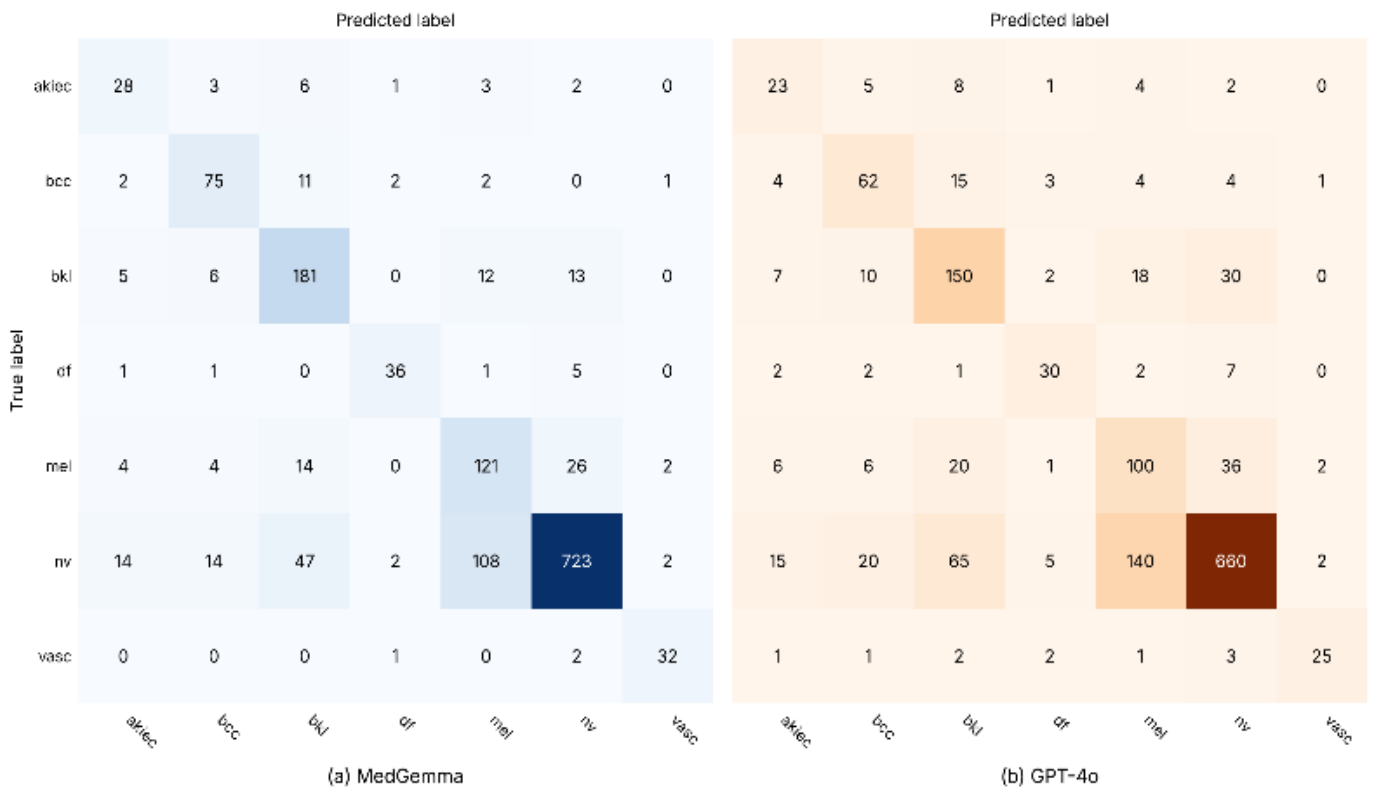

Fig. 3. Confusion Matrix for Skin Cancer Classification by (a) MedGemma and (b) GPT-4o

The result reported at Table V shows that, MedGemma achieved a precision of only 51.9% in AKIEC class, while GPT-4o achieved 39.7%. However, MedGemma demonstrated superior performance for Melanocytic Nevus (NV) class by achieving a precision of 93.8%.

TABLE V. CLASSIFICATION REPORT FOR SKIN CANCER CLASSIFICATION

| Model | Class | Precision (%) | Recall (%) | F1-Score (%) |
|---|---|---|---|---|
| MedGemma | akiec | 51.9 | 65.1 | 57.7 |
| | bcc | 72.8 | 80.6 | 76.5 |
| | bkl | 69.9 | 83.4 | 76.1 |
| | df | 85.7 | 81.8 | 83.7 |
| | mel | 49 | 70.8 | 57.9 |
| | nv | 93.8 | 79.5 | 86 |
| | vasc | 86.5 | 91.4 | 88.9 |
| GPT-4o | akiec | 39.7 | 53.5 | 45.5 |
| | bcc | 58.5 | 66.7 | 62.3 |
| | bkl | 57.5 | 69.1 | 62.8 |
| | df | 68.2 | 68.2 | 68.2 |
| | mel | 37.2 | 58.5 | 45.5 |
| | nv | 88.9 | 72.5 | 79.9 |
| | vasc | 83.3 | 71.4 | 76.9 |

In Alzheimer's disease classification confusion matrix illustrated in Fig. 4, MedGemma correctly predicted 756 non-demented classes, whereas GPT-4o predicted 680 correct non-demented images out of 900 cases. Compared to fine-tuned MedGemma, the classification accuracy of GPT-4o is lower in all cases, and non-demented was also GPT-4o's best class (75.6% recall compared with 84.0% for MedGemma).

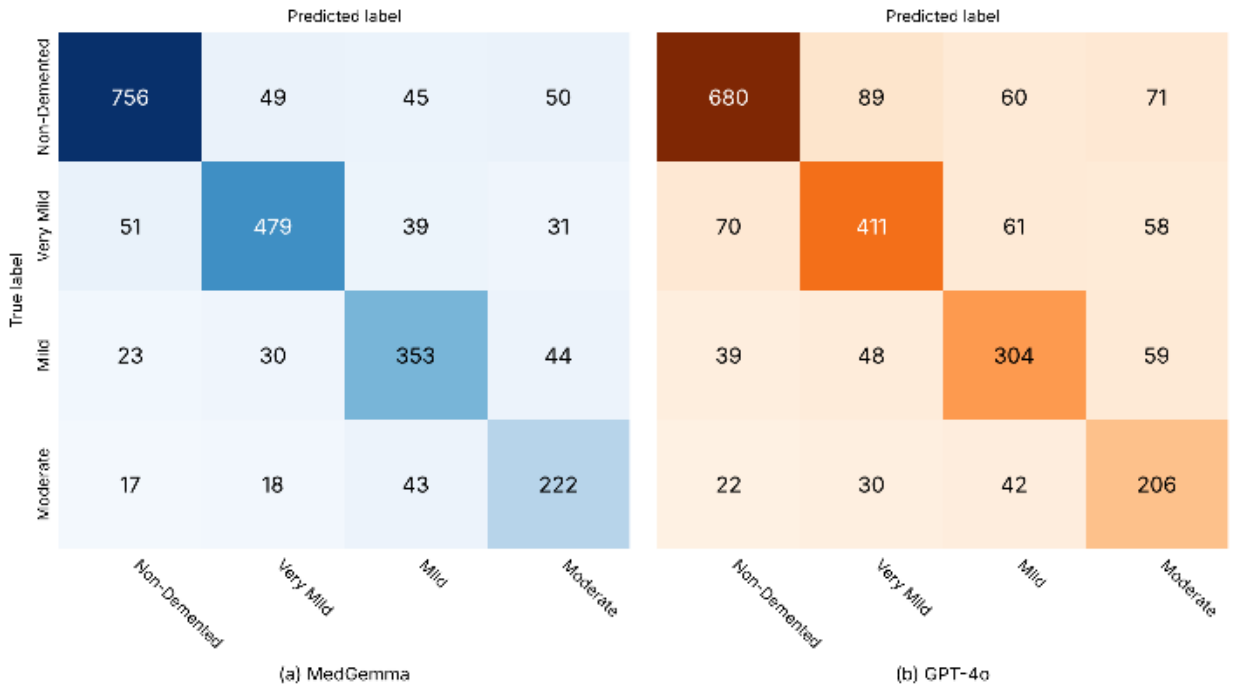

Fig. 4. Confusion Matrix for Alzheimer's Disease Classification by (a) MedGemma and (b) GPT-4o

The comparison in Table VI clearly demonstrates the superior performance of the specialized MedGemma model over the GPT-4o in classifying the four severity stages of dementia. Across all four categories (Non-Demented, Very Mild, Mild, and Moderate), MedGemma achieved consistently higher scores in Precision, Recall, and F1-Score compared to GPT-4o.

TABLE VI. CLASSIFICATION REPORT FOR ALZHEIMER'S DISEASE CLASSIFICATION

| Model | Class | Precision (%) | Recall (%) | F1-Score (%) |
|---|---|---|---|---|
| MedGemma | Non-Demented | 89.3 | 84 | 86.5 |
| | Very Mild | 83.2 | 79.8 | 81.5 |
| | Mild | 73.5 | 78.4 | 75.9 |
| | Moderate | 64 | 74 | 68.6 |
| GPT-4o | Non-Demented | 83.8 | 75.6 | 79.5 |
| | Very Mild | 71.1 | 68.5 | 69.8 |
| | Mild | 65.1 | 67.6 | 66.3 |
| | Moderate | 52.3 | 68.7 | 59.4 |

In Breast Cancer Classification, MedGemma shows strong diagonal dominance in Fig. 5. Table VII shows MedGemma demonstrated balanced and high reliability, achieving F1-Scores of 81.6% for Malignant cases and 83.2% for Benign cases. Its recall (sensitivity) for malignant lesions was 83.3%.

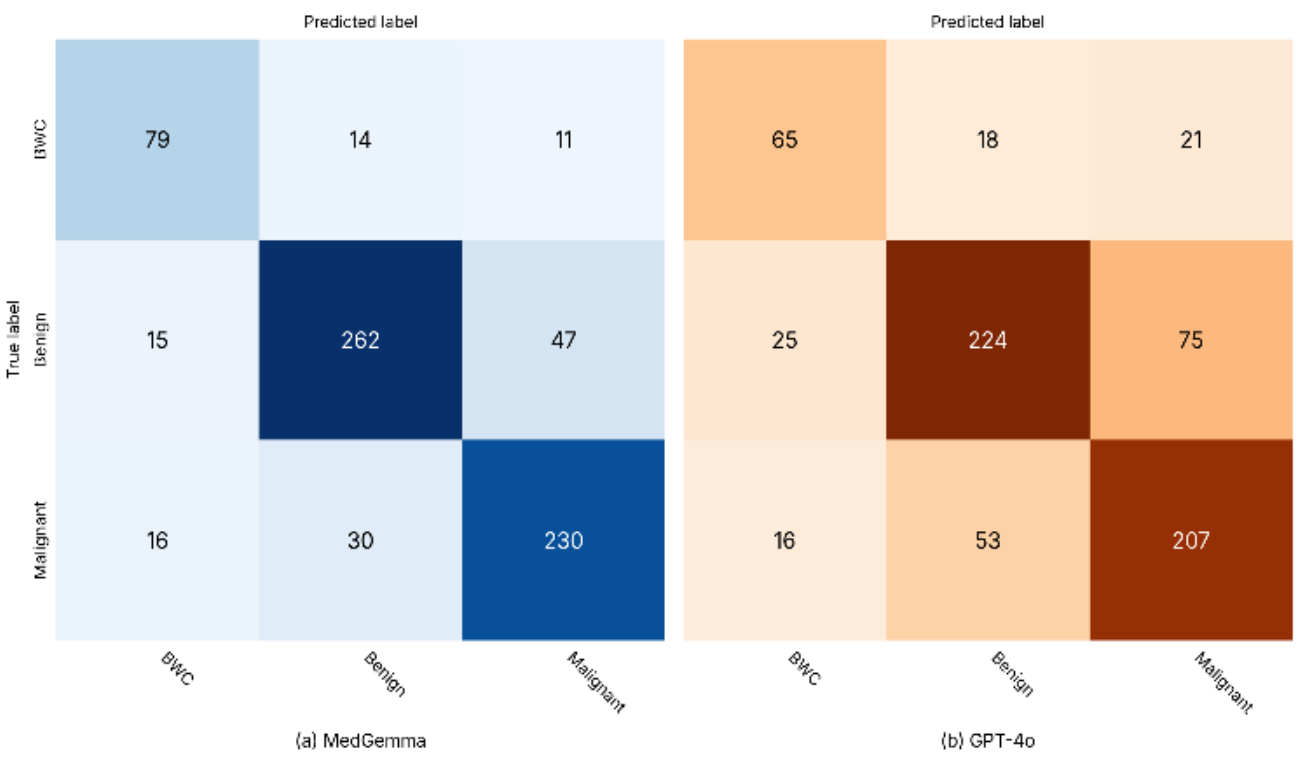

Fig. 5. Confusion Matrix for Breast Cancer Classification by (a) MedGemma and (b) GPT-4o (BWC: benign without callback)

In contrast, GPT-4o reached F1-scores of 71.5% (malignant) and 72.4% (benign), with a malignant recall of 75.0%. Out of 324 cases GPT-4o correctly predicted 224 benign cases. Whereas, MedGemma performed better with 262 correct benign predictions.

TABLE VII. CLASSIFICATION REPORT FOR BREAST CANCER CLASSIFICATION

| Model | Class | Precision (%) | Recall (%) | F1-Score (%) |
|---|---|---|---|---|
| MedGemma | BWC | 71.8 | 76 | 73.8 |
| | Benign | 85.6 | 80.9 | 83.2 |
| | Malignant | 79.9 | 83.3 | 81.6 |
| GPT-4o | BWC | 61.3 | 62.5 | 61.9 |
| | Benign | 75.9 | 69.1 | 72.4 |
| | Malignant | 68.3 | 75 | 71.5 |

In Cardiovascular Disease Classification, MedGemma correctly predicted 230 Myocardial Infarction (MI) from ECG images out of 288 cases while GPT-4o predicted 200 correct MI cases.

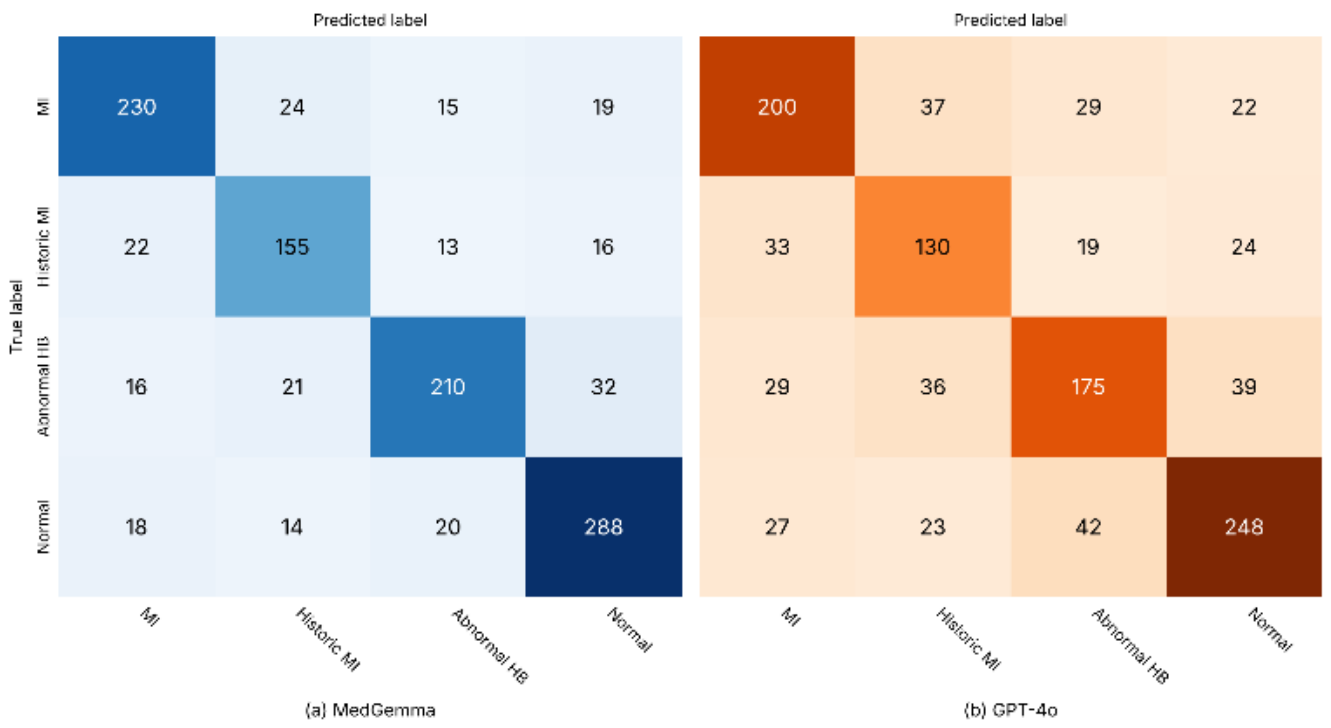

Fig. 6. Confusion Matrix for Cardiovascular Disease Classification by (a) MedGemma and (b) GPT-4o

As reported in Table VIII, MedGemma showed 79.9% recall in MI case and 84.7% recall in Normal case. GPT-4o had only 69.4% recall in MI and 72.9% recall in Normal cases in cardiovascular disease testing set.

TABLE VIII. CLASSIFICATION REPORT FOR CARDIOVASCULAR DISEASE CLASSIFICATION

| Model | Class | Precision (%) | Recall (%) | F1-Score (%) |
|---|---|---|---|---|
| MedGemma | MI | 80.4 | 79.9 | 80.1 |
| | Historic MI | 72.4 | 75.2 | 73.8 |
| | Abnormal HB | 81.4 | 75.3 | 78.2 |
| | Normal | 81.1 | 84.7 | 82.9 |
| GPT-4o | MI | 69.2 | 69.4 | 69.3 |
| | Historic MI | 57.5 | 63.1 | 60.2 |
| | Abnormal HB | 66 | 62.7 | 64.3 |
| | Normal | 74.5 | 72.9 | 73.7 |

As illustrated in Fig. 7, MedGemma achieved 357 correct Pneumonia and 121 correct Normal Case detection whereas GPT-4o achieved only 306 correct Pneumonia detection and 103 correct normal cases detection.

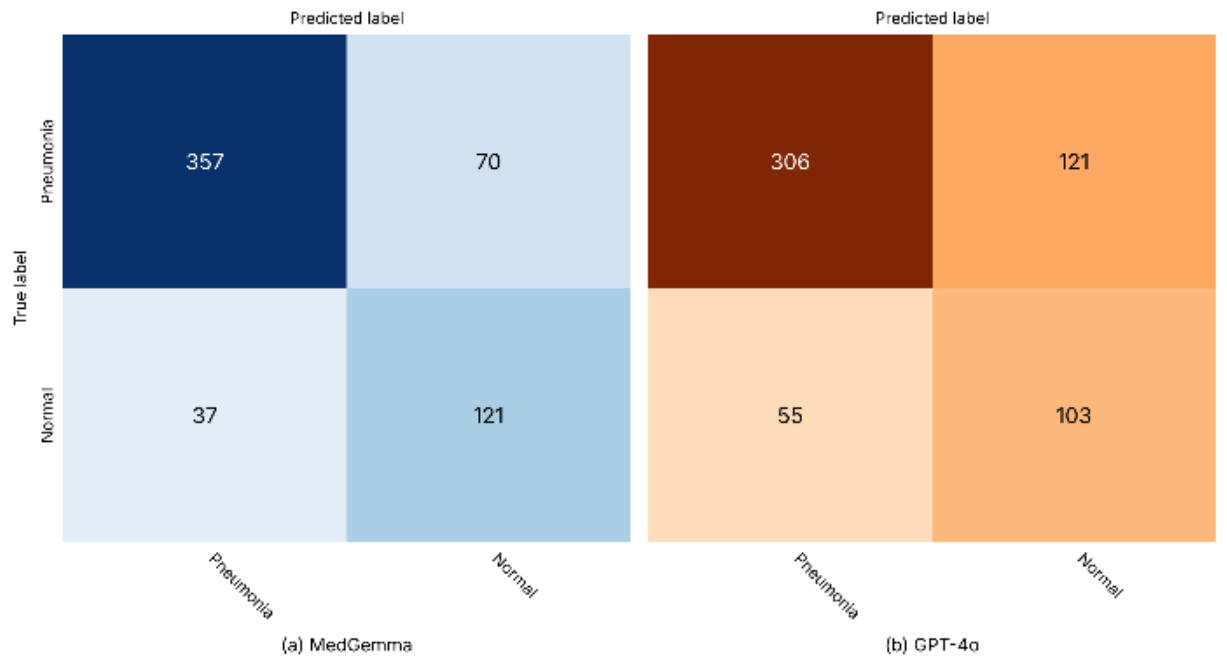

Fig. 7. Confusion Matrix for Pneumonia Classification by (a) MedGemma and (b) GPT-4o

With a recall of 83.6% and an F1-score of 87% for pneumonia (Table IX), MedGemma missed 70 of the 427 pneumonia cases, while GPT-4o missed 121 (recall 71.7%).

TABLE IX. CLASSIFICATION REPORT FOR PNEUMONIA CLASSIFICATION

| Model | Class | Precision (%) | Recall (%) | F1-Score (%) |
|---|---|---|---|---|
| MedGemma | Pneumonia | 90.6 | 83.6 | 87 |
| | Normal | 63.4 | 76.6 | 69.3 |
| GPT-4o | Pneumonia | 84.8 | 71.7 | 77.7 |
| | Normal | 46 | 65.2 | 53.9 |

Across renal images of chronic kidney disease, MedGemma showed strong generalization. As illustrated in Fig. 8, most of the classes are correctly identified. In GPT-4o, a higher number of misclassification is evident.

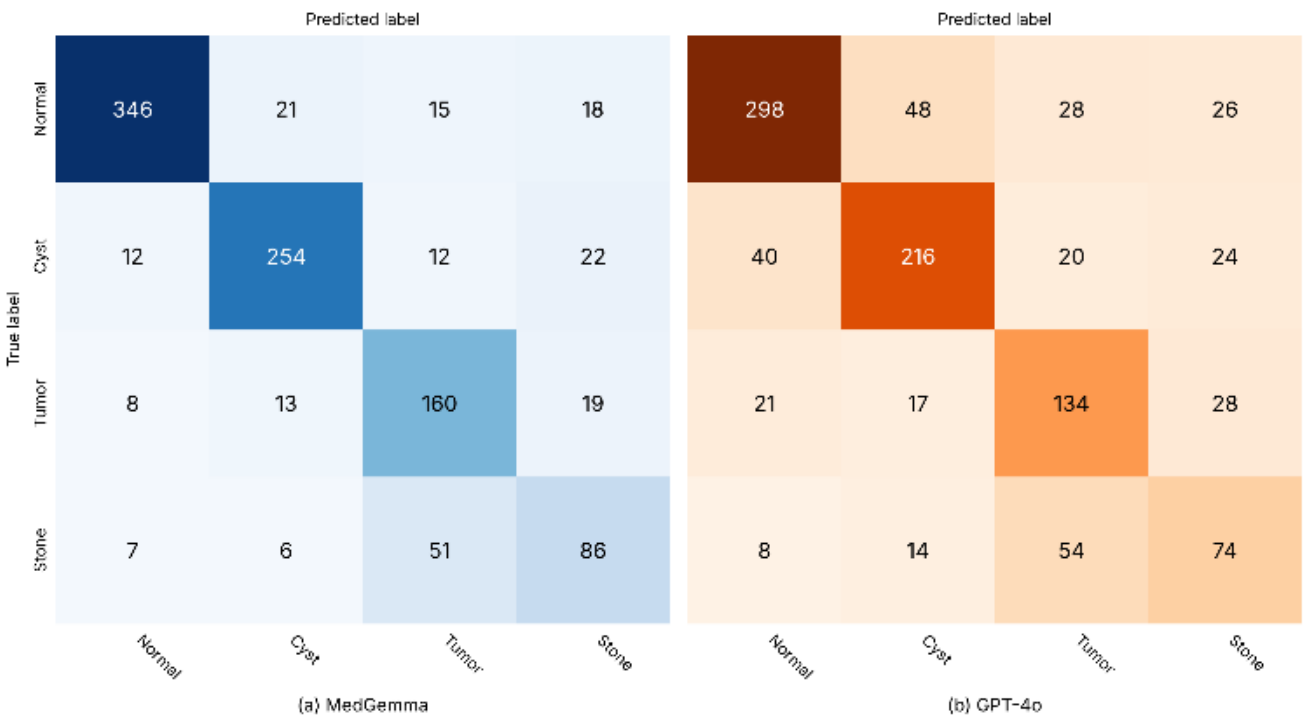

Fig. 8. Confusion Matrix for Chronic Kidney Diseases Classification by (a) MedGemma and (b) GPT-4o

In Table X, MedGemma showed a maximum recall value of 86.5% while GPT-4o's maximum recall value is only 74.5%. MedGemma's weakest kidney class was Stone (57.3% recall): 51 of the 150 stone images were labelled as Tumor, and GPT-4o made the same error on 54 of them.

TABLE X. CLASSIFICATION REPORT FOR CHRONIC KIDNEY DISEASE

| Model | Class | Precision (%) | Recall (%) | F1-Score (%) |
|---|---|---|---|---|
| MedGemma | Normal | 92.8 | 86.5 | 89.5 |
| | Cyst | 86.4 | 84.7 | 85.5 |
| | Tumor | 67.2 | 80 | 73.1 |
| | Stone | 59.3 | 57.3 | 58.3 |
| GPT-4o | Normal | 81.2 | 74.5 | 77.7 |
| | Cyst | 73.2 | 72 | 72.6 |
| | Tumor | 56.8 | 67 | 61.5 |
| | Stone | 48.7 | 49.3 | 49 |

In all cases, MedGemma outperformed GPT-4o by 9.3 to 11.8 percentage points in test accuracy. This substantial gap emphasizes that general AI models like GPT-4o lack the necessary specialized training to handle the subtle, high-resolution features required for accurate medical diagnosis, unlike fine-tuned models such as MedGemma.

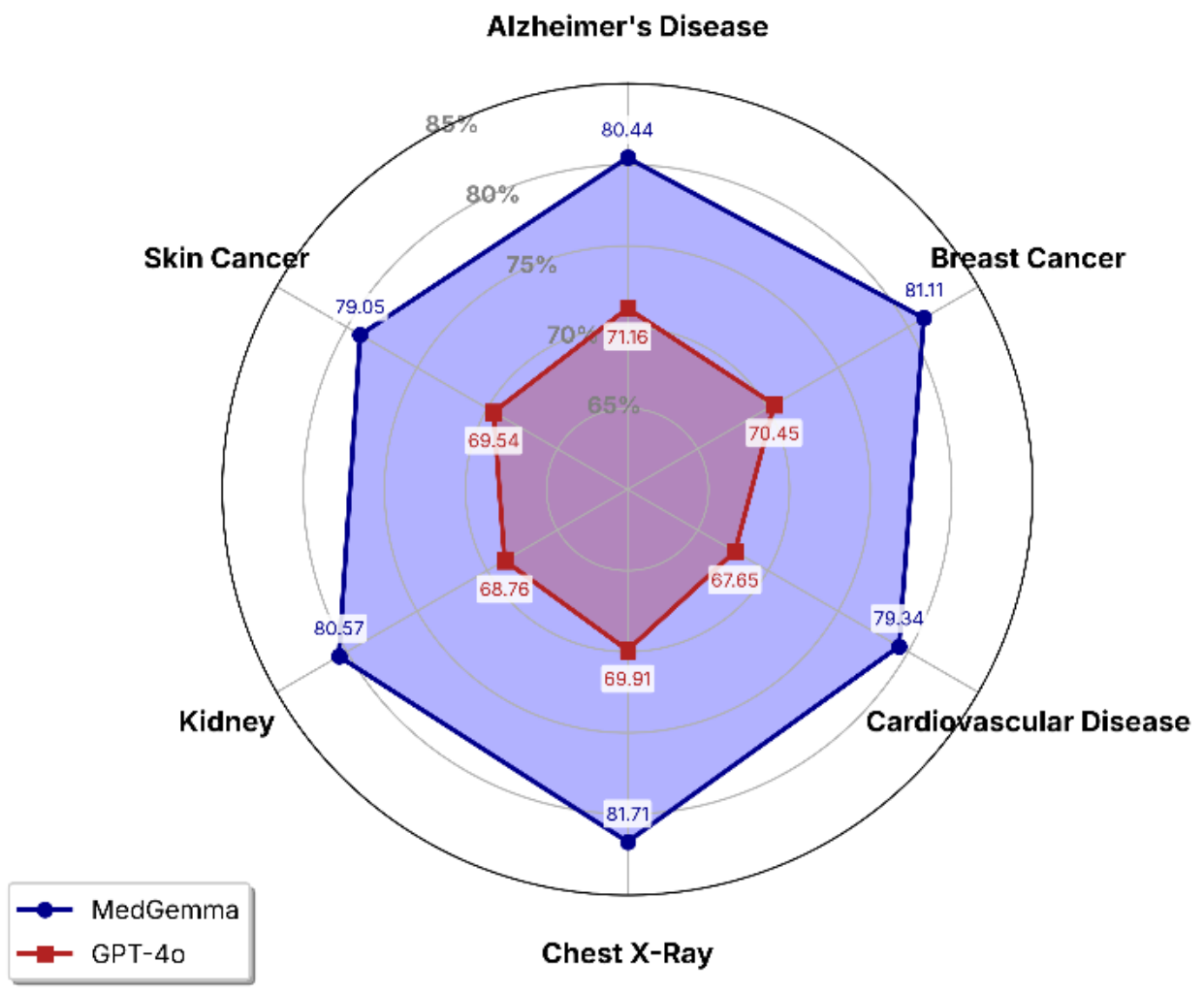

Fig. 9. Radar Chart of MedGemma vs GPT-4o in Test Accuracy across Diseases

The radar chart in Fig. 9 makes it easy to compare the performance of MedGemma-4b-it with GPT-4o, across six medical classification datasets. From a graphical comparison perspective, there is strong evidence that MedGemma demonstrated superior performance than GPT-4o across datasets ranging from diagnosing Breast Cancer to diagnosing chest X-rays. The radar chart illustrates that the MedGemma polygon completely encompasses GPT-4o's polygon, indicating a more robust performance with test accuracy ranging from 79.1% to 81.7%. GPT-4o's test accuracy is lower on every dataset, ranging from 67.7% to 71.2%. Hallucination was not measured directly in this study, so this comparison covers classification errors only.

## V. Findings

The key findings of this work are:

1. Since GPT-4o was used zero-shot without fine-tuning, its accuracy could not match the higher results achieved by the fine-tuned MedGemma.

2. MedGemma's high classification performance on various datasets highlights the value of domain-specific fine-tuning for the field of medical diagnostics.

3. The largest gaps were on chronic kidney disease (11.8 points), pneumonia (11.8 points) and cardiovascular disease (11.7 points), against 10.7 points for breast cancer. This suggests that GPT-4o, as a general model, struggles to analyze the subtle, high-resolution features required for accurate medical image classification.

4. We noticed informally that GPT-4o's answers seemed sensitive to the order in which the image and the prompt were supplied. This was not tested systematically and needs a controlled experiment before it can be treated as a finding.

5. Even with 4 billion parameters, MedGemma reached a modest mean test accuracy of 80.37%. No CNN baseline was trained in this study; a likely reason for the modest accuracy is that MedGemma was fine-tuned using Low-Rank Adaptation (LoRA) on a limited scale due to time and resource constraints.

## VI. Conclusion

This study demonstrates that the MedGemma-4b-it model consistently outperforms GPT-4o across multiple disease classification tasks, achieving a mean accuracy of 80.37% compared to 69.56% for GPT-4o. The performance evaluation between the fine-tuned MedGemma and zero-shot GPT-4o highlights a critical trade-off between domain-specific optimization and general-purpose accessibility. While MedGemma's superior performance is expected considering its exposure to 70% of the disease-specific training data, the comparison remains relevant as GPT-4o serves as a benchmark for high-capacity models trained on massive, heterogeneous corpora. However, despite its broad web-scale knowledge base and ease of use, our findings suggest that GPT-4o's lack of specialized refinement makes it less suitable for high-stakes clinical decision-making compared to models explicitly fine-tuned for medical contexts. Consequently, domain-specific adaptation is essential to enhance the reliability and clinical efficacy of AI-driven diagnostics. Despite limitations regarding dataset size and computational resources, the flexibility and efficiency of MedGemma position it as a promising tool for real-world clinical deployment. Future work should prioritize MedGemma's large-scale clinical validation, ethical and regulatory considerations, and seamless integration into healthcare workflows. Furthermore, advancing hallucination-aware training and evaluation strategies remains essential to ensuring trustworthy, safe, and interpretable AI systems in high-stakes medical settings.

## References

[1] International Diabetes Federation, "IDF Diabetes Atlas," 11th ed., Brussels, Belgium, 2025. [Online]. Available: https://www.diabetesatlas.org

[2] World Health Organization, "Breast Cancer Fact Sheet." 2020. [Online]. Available: https://www.who.int/news-room/fact-sheets/detail/breast-cancer

[3] World Health Organization, "Pneumonia Fact Sheet." 2021. [Online]. Available: https://www.who.int/news-room/fact-sheets/detail/pneumonia

[4] Alzheimer's Association, "2024 Alzheimer's disease facts and figures," *Alzheimer's & Dementia*, vol. 20, no. 5, pp. 3708–3821, Apr. 2024. [Online]. Available: https://pmc.ncbi.nlm.nih.gov/articles/PMC11095490/

[5] World Health Organization, "Cardiovascular Diseases Fact Sheet." Jul. 2025. [Online]. Available: https://www.who.int/news-room/fact-sheets/detail/cardiovascular-diseases-(cvds)

[6] J. Guo et al., "Global, regional, and national burden inequality of chronic kidney disease, 1990–2021: a systematic analysis for the global burden of disease study 2021," *Front. Med., vol. 11, Art. no. 1501175, Jan. 2025.* [Online]. Available: https://www.frontiersin.org/journals/medicine/articles/10.3389/fmed.2024.1501175/full

[7] A. Reshad, V. Nino, and M. Valero, "Deep Learning-Based Detection of Arrhythmia Using ECG Signals – A Comprehensive Review," *Vasc. Health Risk Manag.*, vol. Volume 21, pp. 685–703, Aug. 2025, doi: 10.2147/VHRM.S508620.

[8] K. Singhal *et al.*, "Large Language Models Encode Clinical Knowledge," Dec. 26, 2022, *arXiv*: arXiv:2212.13138. doi: 10.48550/arXiv.2212.13138.

[9] K. Singhal *et al.*, “Towards Expert-Level Medical Question Answering with Large Language Models,” May 16, 2023, *arXiv*: arXiv:2305.09617. doi: 10.48550/arXiv.2305.09617.
[10] H. Takita *et al.*, “A systematic review and meta-analysis of diagnostic performance comparison between generative AI and physicians,” *Npj Digit. Med.*, vol. 8, no. 1, p. 175, Mar. 2025, doi: 10.1038/s41746-025-01543-z.
[11] S. Abbas *et al.*, “THE ROLE OF ARTIFICIAL INTELLIGENCE IN PERSONALIZED MEDICINE AND PREDICTIVE DIAGNOSTICS – A NARRATIVE REVIEW,” *Insights-J. Health Rehabil.*, vol. 3, no. 1 (Health&Allied), pp. 624–631, Feb. 2025, doi: 10.71000/k6cga886.
[12] P. Hager *et al.*, “Evaluation and mitigation of the limitations of large language models in clinical decision-making,” *Nat. Med.*, vol. 30, no. 9, pp. 2613–2622, Sept. 2024, doi: 10.1038/s41591-024-03097-1.
[13] T. Tu *et al.*, “Towards conversational diagnostic artificial intelligence,” *Nature*, vol. 642, no. 8067, pp. 442–450, June 2025, doi: 10.1038/s41586-025-08866-7.
[14] T. B. Brown *et al.*, “Language models are few-shot learners,” in *Proc. 34th Int. Conf. Neural Inf. Process. Syst. (NIPS '20)*, Vancouver, BC, Canada, 2020, Art. no. 159, doi: 10.5555/3495724.3495883.
[15] A. Radford *et al.*, “Language Models are Unsupervised Multitask Learners,” *arXiv e-prints*, 2019, Art. no. 1907.10522, doi: 10.48550/arXiv.1907.10522. [Online]. Available: https://arxiv.org/abs/1907.10522
[16] D. M. Korngiebel and S. D. Mooney, “Considering the possibilities and pitfalls of Generative Pre-trained Transformer 3 (GPT-3) in healthcare delivery,” *npj Digit. Med.*, vol. 4, no. 1, Art. no. 93, 2021, doi: 10.1038/s41746-021-00464-x.
[17] E. Ullah et al., “Challenges and barriers of using large language models (LLM) such as ChatGPT for diagnostic medicine with a focus on digital pathology – a recent scoping review,” *Diagn Pathol*, vol. 19, no. 1, Art. no. 43, 2024, doi: 10.1186/s13000-024-01464-7.
[18] A. Lecler, L. Duron, and P. Soyer, “Revolutionizing radiology with GPT-based models: Current applications, future possibilities and limitations of ChatGPT,” *Diagn. Interv. Imaging*, vol. 104, no. 6, pp. 269–274, 2023, doi: 10.1016/j.diii.2023.02.003.
[19] K. Singhal *et al.*, “Large language models encode clinical knowledge,” *Nature*, vol. 620, no. 7972, pp. 172–180, 2023, doi: 10.1038/s41586-023-06291-2.
[20] J. Lee et al., “BioBERT: a pre-trained biomedical language representation model for biomedical text mining,” *Bioinformatics*, vol. 36, no. 4, pp. 1234–1240, 2019, doi: 10.1093/bioinformatics/btz682.
[21] E. Alsentzer et al., “Publicly Available Clinical BERT Embeddings,” arXiv e-prints, 2019, Art. no. 1904.03323, doi: 10.48550/arXiv.1904.03323. [Online]. Available: https://arxiv.org/abs/1904.03323
[22] A. Sellergren, S. Kazemzadeh, T. Jaroensri, A. Kiraly, M. Traverse, T. Kohlberger, et al., “MedGemma Technical Report,” arXiv e-prints, arXiv:2507.05201 [cs.AI], 2025. [Online]. Available: https://arxiv.org/abs/2507.05201.
[23] OpenAI et al., “GPT-4 Technical Report,” arXiv preprint arXiv:2303.08774, 2024. [Online]. Available: https://arxiv.org/abs/2303.08774.
[24] N. Prentzas, A. Kakas, and C. S. Pattichis, “Explainable AI applications in the Medical Domain: a systematic review,” arXiv e-prints, 2023, Art. no. 2308.05411, doi: 10.48550/arXiv.2308.05411. [Online]. Available: https://arxiv.org/abs/2308.05411
[25] A. Dosovitskiy et al., “An Image is Worth 16x16 Words: Transformers for Image Recognition at Scale,” International Conference on Learning Representations, May 2021, [Online]. Available: https://openreview.net/pdf?id=YicbFdNTTy.
[26] D. Zhou et al., “DeepVIT: Towards Deeper Vision Transformer,” arXiv.org, Mar. 22, 2021. https://arxiv.org/abs/2103.11886.
[27] Y. Xie, “New chapter in pediatric medicine: technological evolution, application, and evaluation system of large language models,” European Journal of Pediatrics, vol. 184, no. 12, p. 809, Dec. 2025, doi: 10.1007/s00431-025-06602-x.
[28] P. Tschandl, C. Rosendahl, and H. Kittler, “The HAM10000 dataset, a large collection of multi-source dermatoscopic images of common pigmented skin lesions,” *Sci Data*, vol. 5, no. 1, Art. no. 180161, 2018, doi: 10.1038/sdata.2018.161.
[29] D. S. Marcus, T. H. Wang, J. Parker, J. G. Csernansky, J. C. Morris, and R. L. Buckner, “Open access series of imaging studies (OASIS): Cross-sectional MRI data in young, middle aged, nondemented, and demented older adults,” Journal of Cognitive Neuroscience, vol. 19, no. 9, pp. 1498–1507, 2007. doi: 10.1162/jocn.2007.19.9.1498. [Online]. Available: https://doi.org/10.1162/jocn.2007.19.9.1498
[30] R. S. Lee, F. Gimenez, A. Hoogi, K. K. Miyake, M. Gorovoy, and D. L. Rubin, “A curated mammography data set for use in computer-aided detection and diagnosis research,” Scientific Data, vol. 4, p. 170177, 2017. doi: 10.1038/sdata.2017.177. [Online]. Available: https://doi.org/10.1038/sdata.2017.177 .
[31] A. H. Khan and M. Hussain, “ECG Images Dataset of Cardiac Patients,” Mendeley Data, V2, 2021. doi: 10.17632/gwbz3fsgp8.2. [Online]. Available: https://doi.org/10.17632/gwbz3fsgp8.2
[32] D. Kermany, K. Zhang, and M. Goldbaum, “Labeled Optical Coherence Tomography (OCT) and Chest X-Ray Images for Classification,” Mendeley Data, V2, 2018, doi: 10.17632/rscbjbr9sj.2.
[33] N. M. Islam, “CT KIDNEY DATASET: Normal-Cyst-Tumor and Stone,” Kaggle, 2021. [Online]. Available: https://www.kaggle.com/datasets/nazmul0087/ct-kidney-dataset-normal-cyst-tumor-and-stone. [Accessed: Oct. 22, 2025].
[34] E. J. Hu et al., “LoRA: Low-Rank Adaptation of Large Language Models,” arXiv e-prints, 2021, Art. no. 2106.09685, doi: 10.48550/arXiv.2106.09685. [Online]. Available: https://arxiv.org/abs/2106.09685
[35] P. Zhou, X. Xie, Z. Lin, and S. Yan, “Towards Understanding Convergence and Generalization of AdamW,” *IEEE Transactions on Pattern Analysis and Machine Intelligence*, vol. 46, no. 9, pp. 6486–6493, Sept. 2024, doi: 10.1109/TPAMI.2024.3382294.
[36] OpenAI et al., “GPT-4o System Card,” arXiv preprint arXiv:2410.21276, 2024. [Online]. Available: https://arxiv.org/abs/2410.21276
[37] Google Health, “Fine-tune MedGemma with Hugging Face,” MedGemma GitHub repository, notebook fine_tune_with_hugging_face.ipynb, 2025. [Online]. Available: https://github.com/google-health/medgemma/blob/main/notebooks/fine_tune_with_hugging_face.ipynb
[38] awsaf49, “CBIS-DDSM: Breast Cancer Image Dataset,” Kaggle. [Online]. Available: https://www.kaggle.com/datasets/awsaf49/cbis-ddsm-breast-cancer-image-dataset. [Accessed: Oct. 27, 2026].